\documentclass{article}
\usepackage{times}
\usepackage{graphicx}
\usepackage{subfigure}
\usepackage{natbib}
\usepackage{algorithm}
\usepackage{algorithmic}
\usepackage{hyperref}
\usepackage{amsmath}

\usepackage[usenames,dvipsnames]{xcolor}

\usepackage[accepted]{icml2018}

\newcommand\blfootnote[1]{%
  \begingroup
  \renewcommand\thefootnote{}\footnote{#1}%
  \addtocounter{footnote}{-1}%
  \endgroup
}

\newcommand\longname{Metric-Agnostic Conditional~}
\newcommand\shortname{MACO~}

\icmltitlerunning{Few-Shot Learning with Metric-Agnostic Conditional Embeddings}
\begin{document}

\twocolumn[
\icmltitle{Few-Shot Learning with Metric-Agnostic Conditional Embeddings}

%\begin{icmlauthorlist}
%\icmlauthor{name}{affiliations}
%\end{icmlauthorlist}

%\icmlaffiliation{to}{University of Torontoland, Torontoland, Canada}
%\icmlaffiliation{goo}{Googol ShallowMind, New London, Michigan, USA}

%\icmlcorrespondingauthor{Cieua Vvvvv}{c.vvvvv@googol.com}
%\icmlcorrespondingauthor{Eee Pppp}{ep@eden.co.uk}

\icmlsetsymbol{equal}{*}

\begin{icmlauthorlist}
\icmlauthor{Nathan Hilliard}{src,equal}
\icmlauthor{Lawrence Phillips}{richland,equal}
\icmlauthor{Scott Howland}{richland}
\icmlauthor{Art\"{e}m Yankov}{src}
\icmlauthor{Courtney D. Corley}{richland}
\icmlauthor{Nathan O. Hodas}{richland}
\end{icmlauthorlist}

\icmlaffiliation{src}{Pacific Northwest National Laboratory Seattle Research Center, Seattle, Washington, USA}
\icmlaffiliation{richland}{Pacific Northwest National Laboratory, Richland, Washington, USA}

\icmlcorrespondingauthor{Nathan Hilliard}{nathan.hilliard@pnnl.gov}

\icmlkeywords{few-shot learning, deep learning, machine learning, computer vision}

\vskip 0.3in
]

%\printAffiliationsAndNotice{}

\begin{abstract}

\blfootnote{${}^*$Equal contribution. ${}^1$Pacific Northwest National Laboratory Seattle Research Center, Seattle, Washington, USA. ${}^2$Pacific Northwest National Laboratory, Richland, Washington, USA. Correspondence to: Nathan Hilliard $<$nathan.hilliard@pnnl.gov$>$.}

Learning high quality class representations from few examples is a key problem in metric-learning approaches to few-shot learning.
To accomplish this, we introduce a novel architecture where class representations are conditioned for each few-shot trial based on a target image.
%We introduce a novel architecture for few-shot learning which achieves state-of-the-art performance on the Caltech-UCSD birds fine-grained classification task.
We also deviate from traditional metric-learning approaches by training a network to perform comparisons between classes rather than relying on a static metric comparison.
%Our architecture is a modification of traditional metric-learning approaches, where a learned network performs the comparison on metric space.
%Importantly, this metric-agnostic approach relies on first conditioning each class representation with the target, or query, image.
This allows the network to decide what aspects of each class are important for the comparison at hand.
We find that this flexible architecture works well in practice, achieving state-of-the-art performance on the Caltech-UCSD birds fine-grained classification task.
%We provide an architecture that permits a conditional metric learning, where the metric space is warped for each task to accommodate minor task-to-task variations common in few-shot learning.
%We draw from the existing literature taking advantage of previous architectures such as relational and prototype networks. Our network is split into four distinct stages which better allows the network to divide the task of representation learning and classification.

%Current few-shot learning models achieve high classification performance using only positive examples from each class. In this work we demonstrate that \emph{negative} examples, images to which the target does not belong, can also be used to improve few-shot learning. We compare our metric learning approach both with and without negative examples against a number of strong metric- and meta-learning baselines. While negative examples improve performance on datasets with broad categories (e.g. miniImageNet), we find that negative examples are particularly useful for fine-grained classification as in the Caltech-UCSD Birds dataset.
%We hypothesize that the utility of negative examples comes from highlighting class boundaries. Using hand-chosen negative examples we demonstrate how class representations are systematically altered based on similarity between negative and positive examples.

\end{abstract}

\section{Introduction}

The goal of few-shot learning is to generalize a classifier's performance to new classes given relatively small amounts of data. Although both adults and children are capable of efficiently making these generalizations~\citep{swingley2010fast,omniglot}, few-shot classification has remained a difficult problem for machine learning algorithms. A key insight from psychology is that human few-shot generalization only occurs when new classes can be understood in the context of old ones~\citep{carey1978acquiring,swingley2010fast}. In essence, the ability to rapidly understand a new category (e.g., a new word) can only be accomplished when the learner already has an idea of what the space of categories looks like. As Carey puts it, ``there must be powerful processes that establish and maintain lexical entries of newly heard words, locating their meanings in some relevant part of semantic space, while the nuanced meaning gets worked out."~\citep{carey2010beyond}.

This notion of semantic, or conceptual, space underlies many approaches to few-shot learning, most directly the set of deep learning models which fall under the umbrella of metric-learning~\citep{kulis2013metric,siamesenets,matchingnets,prototypical}. Metric-learning approaches attempt to solve the few-shot problem by rapidly placing new categories within a learned metric space where classes can be easily separated, most often through a pre-defined distance metric such as Euclidean or cosine distance. These systems have achieved strong performance on many few-shot tasks~\citep{matchingnets,prototypical}, but it is unclear what aspects of their structure are most important for good few-shot generalization.

One important aspect of these models is the relation between learned class representations and the placement of a query image within the metric space. Consider, for instance, a case where a query image shares similarities to a number of few-shot classes based on attributes such as shape or color.
While children, and some neural networks, share a bias to classify based on shape~\citep{landau1988importance,ritter2017cognitive}, we can also recognize the potential that some classifications would require a re-weighting of attribute importance as in the classification of non-solids where shape is less important~\citep{soja1991ontological}.
Accomplishing this feature re-weighting requires an interaction between query and class representations which has not been thoroughly investigated in the literature.

Another area left largely unexplored is whether or not the use of pre-defined metrics, such as cosine or Euclidean distance, is necessary for the strong performance seen in existing metrics-based few-shot learning systems. While \citet{prototypical} explore how switching between different distances affects model performance, it is unclear whether similar, or better, results could be obtained by allowing a parameterized network to perform classification directly.

To explore whether few-shot performance necessarily depends on a true metric space, we introduce a novel few-shot architecture where the metric space comparison is replaced by a learned neural network architecture. This means that our architecture is not a true metric-learning approach, as the output of the model is a softmax probability distribution, and not a true distance or similarity metric. We find, however, that our network performs quite well in practice. We also show that this metric-agnostic aspect is not enough for good performance. The network achieves its best performance only when it is allowed to condition each class representation based on the target, or query, image. Because this final network is both metric-agnostic and creates conditional representations, we refer to it as a \longname network (MACO).

\section{Architecture}
\label{sec:methods}

%The principle of our approach is quite simple: by introducing negative examples into our architecture, our model is able to learn contrastive features to help it differentiate between classes.
%This technique is particularly powerful for fine grained classification tasks where the differences between classes are smaller than broad category classification.
%Even if the negative examples are not derived from a class present in a given $K$-way classification task, it still provides a sharp performance improvement over previous state-of-the-art methods.

%\subsection{Architecture}

%\hilliard{TODO: Need to normalize/clean up the notation in the equations and the text.}

%\hilliard{Capacity plays a big role in training and generalization here. Too deep of a residual and we overfit. If the layers are too narrow, we don't learn effectively--too wide and we overfit. This is particularly applicable in the relational and comparison architectures. Though the diagrams show one depth, it might be worth discussing this trade-off. I believe even just one layer can do a good job (we achieved roughly 63\% with one layer in each block). \\ \\ Another thing is the inclusion of batch norm, we've tried dropout instead but the network was \emph{very} sensitive to this--a difference of roughly 60\% vs 65\% in the dropout rate (with only two dropout layers per block, one before the residual and one after) would cause the network to either not train at all or overfit.}

We describe our network in the traditional fashion used for few-shot learning.
Each few-shot trial is made up of a so-called \emph{query} image, $q$, where the goal of the network is to decide which of $K$ classes the query belongs to. Each class is represented by a set of images and we refer to the entirety of these as the \emph{support set}, $S_K$. For a given class $k$, $S_k$ refers to the images in the support set belonging to $k$. An experiment with $K$ classes and $n$ images per class is referred to as a $K$-way, $n$-shot experiment.

The network is made up of four distinct stages which are trained fully end-to-end. We explicitly separate these components of the model so that individual pieces of the network are not forced to encode multiple complex relationships which might interfere with one another.

\begin{enumerate}

\item \textbf{Feature Stage --} Convolutional architecture to represent images as a single vector. A single set of parameters is used for all images.

\item \textbf{Relational Stage --} Images within a class are compared in a pairwise fashion as in~\citet{relationalnets} resulting in a single vector to represent the class. Parameters are shared across all classes.

\item \textbf{Conditioning Stage --} The query image is used to augment the representation for each class.

\item \textbf{Classifier --} Information is combined across class vectors and a final softmax classification is made based on the query image.
\end{enumerate}

\begin{figure*}[h]
  \centering
  \includegraphics[scale=0.5]{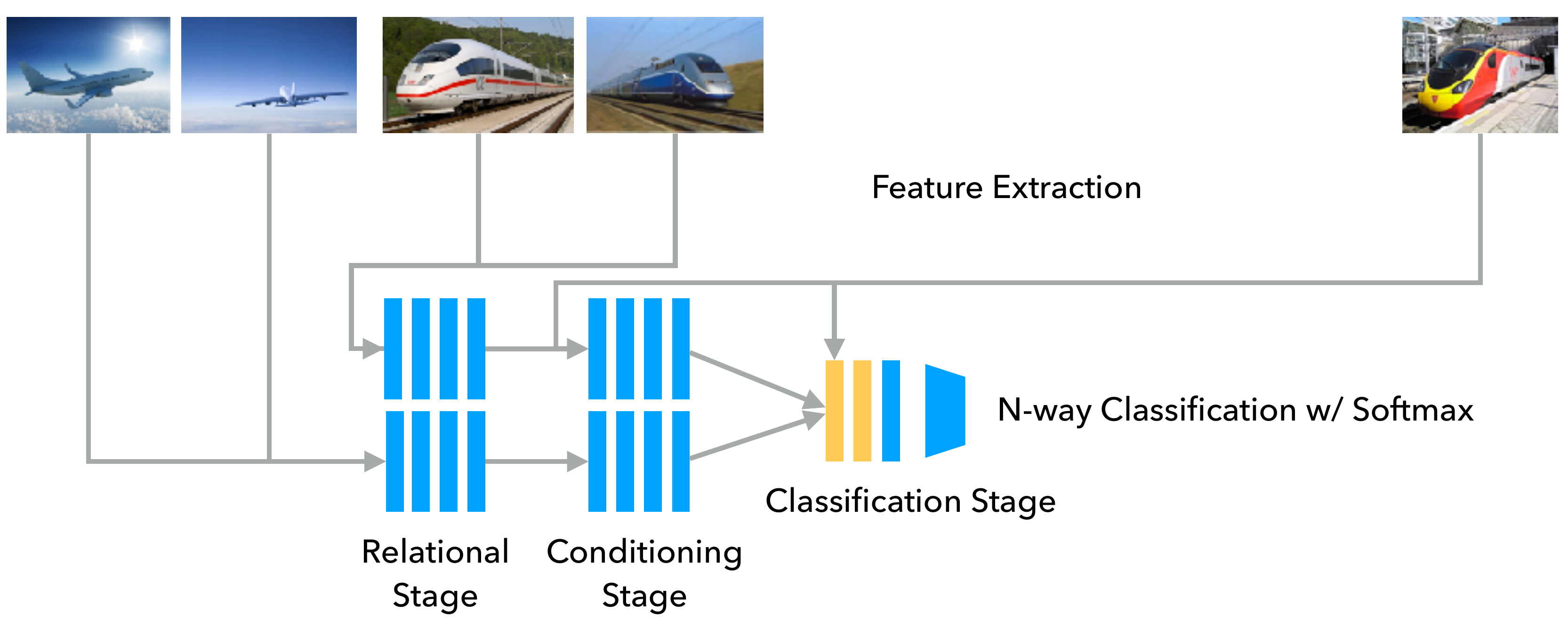}
  %\vspace*{2mm}
  \caption{Full model architecture. Blue rectangles indicate fully connected blocks, while orange rectangles indicate convolutional blocks. The final classification architecture makes use of two 1D convolutional blocks followed by a single fully connected block and dense softmax layer.}
  \label{fig:full_architecture}
\end{figure*}

The full model architecture is detailed in Figure~\ref{fig:full_architecture}. %$q$ represents the query image while $S_k$ represents the set of images belonging to class $k$.
Each rectangle represents a series of blocks, described in more detail below, where blue represents convolutional and orange represents fully connected layers. Note that the query image is incorporated both into the conditioning stage as well as the final classification stage.

\subsection{Feature Stage}
\label{sec:feature}

To extract a feature vector from each image, we use the same convolutional architecture as in~\citet{metalstm}\footnote{We make use of this smaller architecture in order to more fairly compare against baseline models in the literature. Initial experiments indicate the model also works well with a larger, pre-trained ResNet architecture.}. The network is made up of four convolutional blocks where each block begins with a 2D convolutional layer with a $3\times3$ kernel and filter size of 32. Each convolutional layer is followed by batch normalization, an ELU activation~\cite{elu}, and a $2\times2$ max pooling layer\footnote{Initial experiments with dropout as a regularizer showed poor convergence during training and are therefore excluded.}. After the fourth convolutional block, a linear layer produces a vector of size 800 to represent the image. This feature architecture is used with the same parameters for all images in each few-shot trial, regardless of whether the image is a query or from the support set. This encourages the network to use the feature stage to learn generic visual features which are useful regardless of the status of the image in the few-shot trial.

\subsection{Relational Stage}
\label{sec:relational}

\begin{figure}
  \centering
  \includegraphics[scale=0.6]{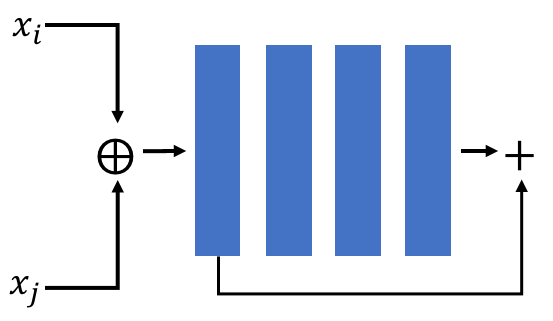}
  \vspace*{2mm}
  \caption{Relational network $g$. $x_i$ and $x_j$ are image vectors of size 800 processed by the convolutional feature extractor. Each block represents a fully connected layer, batch normalization, and ELU non-linearity. The final output of the block is the summation of the first and final block outputs. A final class output is created by averaging the output of each image comparison.}
  \label{fig:relational}
\end{figure}

A key problem in few-shot learning is how to efficiently learn class representations from a small set of images and previous approaches make use of a variety of techniques. Matching networks take advantage of the set-to-set framework with an attention kernel over images in $S_k$~\citep{vinyals2015order,matchingnets}. Prototypical networks simply use a feature stage to embed images into the metric space and use their average to represent the class.

To combine information across images in a class, we turn to a different set-based network architecture, relational networks~\citep{relationalnets}. Relational networks take a set as input and output a single representation that is order insensitive.
While the original relational network was intended to process multiple areas within a single image, in the case of few-shot learning, we can treat the images of a single class in the support set, $S_k$, as input to a relational network because their ordering is irrelevant.
Comparing to prototypical networks, this allows us to learn a more complex relationship between the images in $S_k$ and the vector representation for class $k$.
This method also allows us to avoid imposing an arbitrary ordering onto $S_k$ as in the case of matching networks with full context embeddings~\citep{matchingnets}.

%As \citet{prototypical} make note of, the more complex, Full Context Embedding version of matching networks also forces $S_k$ into an arbitrary ordering which we are able to avoid through a relational network.

In the original formulation, a relational network is a network that takes in two items at a time and produces a single vector. Pair-wise comparisons are made using the same network for every pair of items within the input set. For a few-shot class with $n$ images, this results in $\binom{n}{2}$ comparisons\footnote{Although this scales quadratically with $n$, we note that a fixed number of sampled comparisons could also be used as an approximation in cases where full calculations would be problematic.}. Relational networks then combine information from these comparisons using a summation. We differ in that our relational stage makes use of an average following the approach of~\citet{automl}. This has the added benefit that the average is invariant to the number of images per class. In cases where $n$ is always the same, as in our experimental results, this method differs only in terms of the scale of the output.

We formalize this as a function in Equation~\ref{eqn:R} where $g$ is a relational network with parameters $\rho$ and $n$ is the number of images in $S_k$. Note that this produces a vector for a single class only. The process is repeated for all $K$ classes to produce $K$ class vectors.
%Parameters for $g$ are shared across all positive classes while a separate set of parameters are used for the class of negative examples. Initial experiments demonstrated that separate parameters were necessary, possibly because the types relations within a set of negative examples are quite different from those of a semantically coherent positive example class.

%the unique combinations in a set $S$ with $n$ examples $S_n$ with a set of parameters $\rho$ to our relational network $g$.

\begin{equation}
\label{eqn:R}
R_\rho(S_k) = \frac{1}{\binom{n}{2}} \sum\limits_{(x_i, x_j) \in S_k} g_\rho(x_i, x_j), i \neq j
%R(S_n, \rho) = \frac{1}{\binom{n}{2}} \sum\limits_{S_i, S_j}^{\binom{S_n}{2}} g_{\rho}(S_i, S_j), i \neq j
\end{equation}

As in the original relational network paper~\citep{relationalnets}, we parameterize $g$ as a network with fully connected layers.
The network is structured similarly to the feature extractor, but using fully connected instead of convolutional blocks. As before, within a block each fully connected layer is followed by batch normalization and an ELU activation. Fully connected layers have dimension 128. We also make use of a skip connection which links the output of the first and final fully connected blocks. Skip, or residual, connections are additional connections made between two layers in a network that ``skip" over one or more intermediary layers and were an important step in efficiently training very deep networks~\cite{skipconnections}. We connect the first and final blocks by summing their individual outputs. This allows later layers of the network to focus on processing information which is not fully captured by the first layer.% \hodas{We should have a Hodas Stinis citation for this shortly}

We treat the number of blocks within the relational stage as a hyperparameter which can be modified based on the complexity of the modeling task. Unless otherwise noted, we make use of 4 fully connected blocks. This relational architecture is used for every pair-wise comparison of images within a few-shot category with the final output of the relational stage being the average output across all comparisons. The output vectors can be thought of as a class embedding into a 128-dimensional embedding space. %This is repeated for every category in the $K$-way comparison so that a total of $K$ vectors are passed onto the next stage.

\subsection{Conditioning Stage}
\label{sec:condition}

\begin{figure}
  \centering
  \includegraphics[scale=0.6]{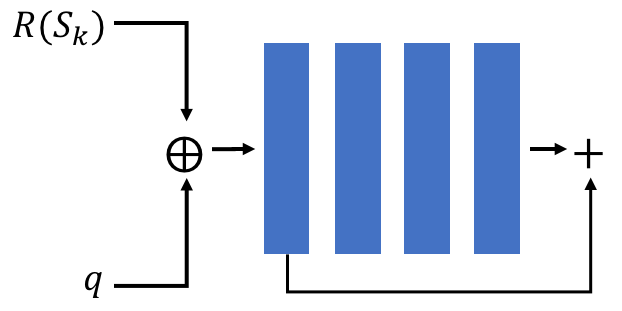}
  \vspace*{2mm}
  \caption{Conditioning network. The layer level design is the same as in Figure~\ref{fig:relational}.}
  \label{fig:comparison}
\end{figure}

Because the relational stage outputs a single vector for each few-shot category, we could simple pass their output directly to the final classification, which would follow the method used in prototypical networks~\cite{prototypical}. Instead, we introduce an intermediate stage where the network has a chance to use information about the query image to condition each individual class representation. We therefore refer to this as our conditioning stage. Conceptually, the output of this stage is a modification of the original class embedding which better takes into account which features of the class might be most relevant for a few-shot trial.

While $R(S_k)$ learns to represent the similarities between images in class $S_k$, we would also like to learn class vectors that take into account the query image, $q$. Doing so gives the network flexibility in learning what aspects of an image class might be relevant to a particular query. For instance, although the relational stage must encode as many features of the class as might be relevant, our conditioning stage might be able to specifically encode a similarity of color or dissimilarity in shape between the class and query. To achieve this, we concatenate $R(S_k)$ and $q$ for each class in $S$ and then allow each to be separately processed by the rest of the conditioning architecture.

The conditioning network is described in Equation~\ref{eqn:C} where $h_\gamma$ is a neural network $h$ parameterized by $\gamma$:

\begin{equation}
\label{eqn:C}
C(S_k, q) = h_\gamma(R_\theta(S_k), q)
\end{equation}

$h$ is structured similarly to the relational network $g$ with a series of fully connected blocks with batch normalization and ELU activations. Unless otherwise mentioned, we make use of 4 blocks in the conditioning network. A skip connection again sums the output of the first and final blocks. $h$ differs only in that its input is the concatenation of two vectors, $R_\theta(S_k)$ and $q$. Fully connected layers again have a dimension of 128, resulting in a conditioned embedding for each class within a 128-dimensional space.

%The comparison block takes the class and anticlass vectors described in section~\ref{sec:relational} as input along with the query image $q$.
%We treat this stage as a mechanism to compress the outputs of $R_\theta(S_k)$ and $R_\phi(A)$ into a single shared representation including $q$.
%Put simply, we take the concatenation of each of those three elements and pass them through a fully connected block similar to that in the previous section.

This structure enables us to produce a single conditioned vector describing the group of images in the context of the query image, allowing the network to adapt its class representations for the given query.
This representation is produced for each class in a traditional $K$-way problem structure, producing $K$ corresponding conditioned vectors.
By updating the class representation in a separate block rather than in the final classification stage we allow the model to separate the problem of understanding the context in a particular experimental trial from the problem of choosing which class the query image belongs to.

To ensure that this portion of the network is being utilized as intended, we also consider a modification of the algorithm where the input to the conditioning stage is simply $R(S_k)$. This removes the ability of the model to condition class representations on the query, but largely retains the additional number of parameters added by the stage. We refer to this as the Metric-Agnostic without conditioning model (MA w/o cond.).

\subsection{Classification Stage}

%\hilliard{The 1D convs here were just sort of off-the-cuff like ``Well lets just use these!" based on prior experience + a fully convolutional output lets us be size invariant in Sharkzor--in our most recent experiments though you really do need a fully connected output layer otherwise the network is difficult to train. We had originally used 1D dilated convs across all of the comparison vectors, but recent results were obtained with just a regular 1D convolution with a kernel size of 3. We haven't tried using a recurrent network here, but it could work--in previous few-shot architectures we used we did find that GRUs/LSTMs combining these vectors would actually just cause the network to memorize everything. Even so, it may work now and might be worth an attempt for comparisons sake.}

\begin{figure}
  \centering
  \includegraphics[scale=0.6]{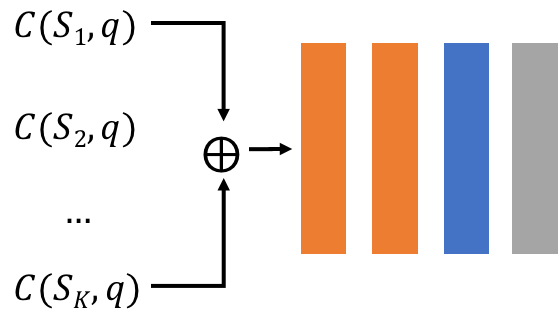}
  \vspace*{2mm}
  \caption{Final classification stage. The first two blocks are one dimensional convolutions with batch normalization and a non-linearity. The flattened output of the last convolution is then fed into a fully connected layer with batch normalization and a non-linearity prior to the final $K$-way output layer with a softmax activation.}
  \label{fig:output}
\end{figure}

Now that we have a vector representing each class in $S_K$, it would be possible to simply embed $q$ within this space and use a pre-defined metric, e.g. Euclidean distance, in order to perform classification. Instead, we opt to replace this with a parameterized neural network which takes in the representations of the support set and query and creates a softmax classification. We note that this classification architecture should not be thought of as a true metric, since it simply learns to point towards the correct class and does not necessarily satisfy all the properties of an actual metric. Notably, the query image is never embedded within the same space as the support set. %On a conceptual level, however, the role it plays is similar.

One goal for the classification architecture is that it should combine information across all five classes in order to make its final decision, rather than an individual decision being made for each class in isolation. To accomplish this we make use of a convolutional architecture without padding that learns to combine information across classes in an order agnostic manner. The input to the classification layer is $n$ 128-dimensional class vectors, which we pass into a 1D convolutional block with a kernel size of 3, filter size 128, batch normalization, and an ELU non-linearity. A second 1D convolutional block of the same specifications reduces all information about a 5-way comparison into a single 128-dimensional vector. A fully connected block of size 128, again with batch norm and an ELU activation performs a final non-linear operation before passing the vector to the final dense softmax layer.

Because the order of classes is randomized for each trial, this final softmax learns to point to the most similar class regardless of its arbitrary ordering.

%The final block of our network is designed to ingest $K$ different conditioned vectors from the previous conditioning stage.
%The goal of this stage is to make sure that the representations from each class can be compared against the query image {\emph as well as against one another} to ensure the final classification is made with as much information as possible.
%Each class vector is concatenated with the features extracted with the query image to produce a single vector capturing both features specific to the query image as well as a representation of the class that is explicitly conditioned on the query (thanks to the conditioning stage).
%We then apply two blocks where each is made up of a one-dimensional convolution with batch normalization and an ELU non-linearity. Each convolutional layer has kernel size of 3 and 64 filters. For our five-way experiments, this ensures that the final convolutional output takes into account information from each of the five classes.
%Finally, the output of the convolutional blocks is flattened and fed to a single fully connected layer of dimension 128 before proceeding to a final $K$-way classification using a fully connected linear layer.

\section{Related Work}
% Experimental Baselines:
%	MAML
%	Meta LSTM
%	Matching Nets
%	Prototypical Nets

%Although there has been a great deal of work on few-shot learning with neural networks, all previous studies have investigated only the case where a class must be inferred from positive examples alone.
% Comparison to metric learning

Our work draws on a number of previous approaches to few-shot learning, predominantly those referred to under the umbrella of metric-learning~\citep{kulis2013metric,matchingnets,prototypical}. The goal of such approaches is to embed the input into a vector space where a simple distance function can be used for classification. Our work differs from traditional metric-learning in that we allow a neural network (our classification stage) to learn both the embedding space and the comparison metric, rather than using a static distance function such as cosine or Euclidean distance.

Architecturally, we take inspiration from methods such as Siamese~\citep{siamesenets} and relational networks~\citep{relationalnets}. Similar to the Siamese network approach, we apply a single network to process images from all classes. Once features for each image have been processed, we make use of the relational network approach of matching networks. Although~\citet{relationalnets} create a single representation by summing across all images in $S_k$, the use of summation limits the model to cases where the number of object comparisons is always identical. Although this is the case in our experiments, we choose to follow~\citet{automl} in averaging across image comparisons, avoiding this fundamental limitation of the sum-based relational net.

Our model also differs substantially from previous attempts in the way it combines information about exemplar images in each class and the query image. Matching networks~\citep{matchingnets} learn to embed individual positive examples by taking into account the full support set $S$, which they label \emph{full context embeddings}. Prototypical networks~\citep{prototypical} can achieve better results by making careful architecture choices but represent each class based only on its own images. This is similar to our relational stage which creates a class vector taking into account only the images within the class. Instead of using the query only at the very end, as is Matching and Prototypical networks, we introduce an intermediary conditioning stage where positive class vectors and the query image can be combined to create an updated class vector, warping the metric space to accommodate the task at hand.

As noted previously, one of the largest differences between our approach and many other metric-learning models is that we do not make use of a pre-defined distance metric. Whereas \citet{matchingnets} make use of cosine distance and \citet{prototypical} make use of Euclidean distance, we allow our network to make use of a classification stage that performs a similar function, but which has learnable parameters. While our parameterized classification stage does not calculate a formal similarity metric per se, it does offer increased flexibility in the modeling task and we find that it performs well in practice.
%For example, the class vectors of a prototypical network must be learned without reference to the query image while still placing themselves within a metric space which aligns with possible query images. Each class will differ from the query in a number of separate features, but

% Mention of meta learning
An alternative approach to few-shot learning is known as meta-learning. Models such as MAML~\cite{maml} and Meta-LSTM~\cite{metalstm} can be thought of as learned optimizers which train a new network to make few-shot decisions for every few-shot trial. Meta-LSTM accomplishes this by incorporating a robust external memory, which in the context of few-shot learning can be used to store information about previously seen classes. The MAML architecture instead frames the problem as one involving two separate, but cooperative, networks. A high-level network, the \emph{meta-learner}, learns to adapt the weights of a low-level network, the \emph{learner}, which makes actual task decisions. For every few-shot learning example, the learner is initialized and then trained for a short duration by the meta-learner with error backpropagated from the learner on to the meta-learner.
Although our approach differs quite substantially from these meta-learning models, they represent some of the currently strongest baselines in few-shot learning and for this reason we compare our model against them.

\section{Experiments}
\label{sec:experiments}

\subsection{Experimental Design}

For each of our experiments we make use of the same architecture and training hyperparameters. Architectural details are given in Section~\ref{sec:methods}. All networks are trained using in Keras using the Nadam optimizer with a learning rate of 0.001. Our models trained best using LeCun normal initialization~\citep{lecun1998efficient} for all fully connected layers and glorot normal~\citep{glorot2010understanding} for convolutional layers. We train each model for 50 epochs with 60,000 few-shot trials per epoch and a batch size of 32. In all cases we evaluate the model with the highest validation accuracy.

Because the network sees the query image twice (by design), the network is prone to overfitting by memorizing the relationships between images in the training set. It is important to prevent the network from memorizing the small sets of images that make up each training class. To deal with these issues, we explored a number of regularization techniques but settled on using only data augmentation. For training we began with an aggressive data augmentation scheme that included randomized rotation, translation, zooming, and horizontal flipping. We found that, in practice, this prevents the models from overfitting without the need for additional regularization such as dropout.

% We additionally use common, fixed values for other experimental parameters that affect the amount of data that the model is given across all baselines.
% Specifically, each method uses a batch size of 32 and trains for a total of 60,000 iterations.

%Our mechanism for negative example selection utilizes a random sampling technique to demonstrate average performance of our approach.
%\hilliard{Maybe good to include a smarter sampling technique that restricts random sampling to be only among classes present in the $K$-way, as an upper bound.}
%The only guarantee that our random sampling approach makes is that the negative examples be derived from classes that are strictly different from the reference image. %\lap{Might want to explicitly go over how this works for test examples}
%For validation and test examples we follow the same procedure but restrict the negative examples to be sampled from the classes for validation and test, respectively.

We include baselines from other well known techniques in this field including Meta-LSTM, MAML, matching networks and prototypical networks\footnote{Baseline implementations were used from:\\Meta-LSTM \& matching networks: \url{https://github.com/twitter/meta-learning-lstm}\\MAML: \url{https://github.com/cbfinn/maml}.\\Prototypical networks: \url{https://github.com/jakesnell/prototypical-networks}
}. All baseline models were evaluated on the same train/val/test splits as the \shortname networks in order to ensure equivalency among the results.
%In order to make sure assess the importance of negative examples, we further implement a version of our model with only positive examples. The positive-only model is identical to that described in Section~\ref{sec:methods} except that the negative example class vector is removed from the comparison stage.

\subsection{Caltech-UCSD Birds}
%\documentclass[../icml18.tex]{subfiles}
%\begin{document}
\begin{table*}[h]
	\centering
	\begin{tabular}{ l  l  l  }
		\hline
		\bf{Model}                           & \bf{1-shot}               & \bf{5-shot}          \\ \hline
		%\bf{Pixel-nearest-neighbor} & 24.25\%                  & 27.28\%    \\
		\bf{Matching Network}                   & 29.34\% & 35.48\% \\
		\bf{Matching Network (FCE)}             & 49.34\%              & 59.31\%     \\
		\bf{Prototypical Network}  & 45.27\% & 34.35\% \\
		\bf{Meta-Learner LSTM}              & 40.43\%     & 49.65\% \\
		\bf{MAML}							& 38.43\%	& 59.15\% \\ \hline
		\bf{\shortname}                          & \bf{60.76}\%                     & \bf{74.96}\%   \\
		\bf{MA w/o cond.}                          & 55.86\%                     & 69.49\%   \\ \hline
	\end{tabular}
	\caption{Average test set classification accuracy on Caltech-UCSD Birds.}
  \label{bird-results}
\end{table*}
%\end{document}

We first look at a fine-grained classification task in the Caltech-UCSD Birds 200 (CUB-200) dataset. This dataset includes 200 fine-grained categories of birds which we randomly divide into 100 for training, 50 for validation, and 50 for testing. Each image is again resized to 84x84 pixels and put through the data augmentation process described previously to reduce overfitting. Initial experiments indicated that the deeper networks, as used on our other datasets, generalized poorly even with data augmentation. To combat this, we reduced the depth of the relational and conditioning stage blocks from 4 to 2.
We present our results on this dataset in Table~\ref{bird-results}.

%We additionally propose the inclusion of the Caltech-UCSD Birds 200 (CUB-200) dataset as another fine-grained classification task.
%This dataset consists of 200 species of birds which we randomly sample from, shuffle, and divide up into 64 for training, 16 for validation, and 20 for testing in order to parallel the setup done in the \emph{mini}ImageNet and \emph{mini}DogsNet experiments.

We find that our model is able to easily outperform the previous state-of-the-art. For the 5-shot experiments, we are able to achieve 74.96\% test accuracy, over 15 percentage points higher than the best performing baseline, matching networks. As expected, performance is much worse in the 1-shot case, with performance at 60.76\%, but is still approximately 11 percentage points higher than the matching networks baseline (49.34\%).

To understand what leads to this level of performance, we investigated whether or not the inclusion of the comparison stage was necessary. In order to keep model parameters roughly comparable, we leave in the conditioning stage but remove the inclusion of the query image. We deem this our Metric-agnostic network without conditioning. The important aspect of this network is that information about the query image can only be included at the final classification stage, which is also where information about all classes in the support set is made. We find that the full \shortname network is able to achieve much higher performance, gaining approximately 5 percentage points accuracy on both 1- and 5-shot tasks.

\begin{figure}
  \centering
  \includegraphics[width=0.5\textwidth]{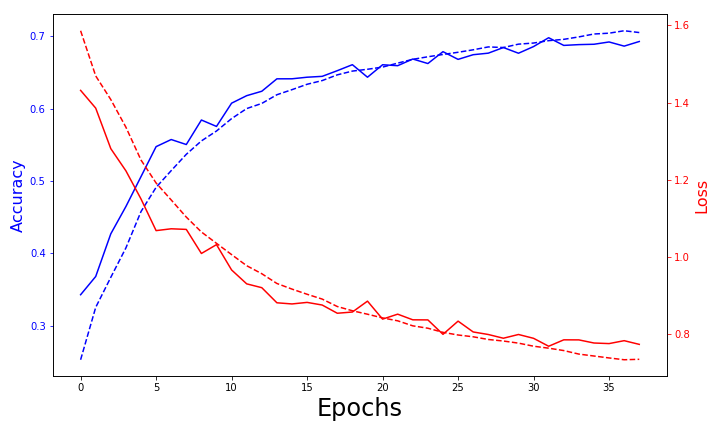}
  %\vspace*{2mm}
  \caption{Loss and accuracy across epochs for 5-shot, 5-way experiments on the Caltech-UCSD Birds dataset. Epochs are arbitrarily defined as 60k iterations. Training loss and accuracy are represented as dashed lines, while validation scores are solid.}
  \label{fig:epochs}
\end{figure}

In Figure \ref{fig:epochs} we show the how model loss and accuracy change over the course of training for the 5-shot task. In blue we represent the accuracy of the model with accuracy on the training set as a dashed line and the validation accuracy as a solid. Model loss is represented as the curve in red. We note that validation accuracy and loss starts off much better because of the aggressive data augmentation which takes place for the training set.

\subsection{\emph{mini}ImageNet}
%\documentclass[../icml18.tex]{subfiles}
%\begin{document}
\begin{table*}[h]
	\centering
	\begin{tabular}{l l l}
		\hline
		\bf{Model}                           & \bf{1-shot}               & \bf{5-shot}          \\ \hline
		%\bf{Pixel-nearest-neighbor} & 22.93\%                     & 24.53\%   \\
		\bf{Matching Network}                   & 43.74\%  & 52.90\% \\
		\bf{Matching Network (FCE)}             & 45.91\%   & 57.66\%      \\
		%\bf{Prototypical} & 0.00\% & 0.00\% \\
		\bf{Meta-Learner LSTM}              & \bf{49.26}\%      & 59.59\% \\
		\bf{MAML}							& 32.05\%   & \bf{61.55}\%				 \\ \hline
		\bf{\shortname}                          &  41.09\%                    &   58.32\%              \\ \hline
	\end{tabular}
	\caption{Average test set classification accuracy on \emph{mini}ImageNet.}
  \label{mini-imagenet-results}
\end{table*}
%\end{document}

To test the ability of our architecture to learn relatively broad categories, we evaluate the model on \emph{mini}ImageNet.
The \emph{mini}ImageNet task, originally defined in \citet{matchingnets}, uses a random set of 100 classes from the ILSVRC ImageNet dataset. We used the same class splits used in \citet{metalstm} with 64 for training, 16 for validation, and the remaining 20 classes as a test set. The original ImageNet images are downsampled to a smaller 84 $\times$ 84 resolution. When generating few-shot trails, we randomly sample our own images from the 600 images provided in each class.

%The goal of the task is to train a few-shot classification network to identify which class a given reference image belongs to.
%We retrain an ImageNet classifier on a subset of the total dataset that does not include classes present in the validation or test sets of this task and fine tune.
%In general, we consider this to be a more broad category classification task since each class is not necessarily directly related, e.g. airplanes versus cats.

%Because the categories in \emph{mini}ImageNet are randomly sampled from a set of relatively broad classes, we do not expect that negative examples will be especially informative for any given trial.
We report our results in conjunction with baselines in Table~\ref{mini-imagenet-results} with the best performing model for a given task identified in bold. We conduct 5-way experiments with both 1-shot and 5-shot trials. Of the baseline models we find that meta-learning approaches are most successful, with Meta-LSTM achieving 49.26\% on the 1-shot trials and MAML achieving 61.55\% on the 5-shot.

Our \shortname network achieves a 5-shot test accuracy of 58.32\%, higher than the matching network baseline but somewhat below the two meta-learning algorithms. We perform more poorly on the 1-shot case with only 41.09\% accuracy.
%These results indicate that the \shortname network is capable of learning the kinds of broad classes which make up the \emph{mini}ImageNet dataset.

\subsection{\emph{mini}DogsNet}
%\documentclass[../icml18.tex]{subfiles}
%\begin{document}
\begin{table*}[h]
	\centering
	\begin{tabular}{l l l}
		\hline
		\bf{Model}                           & \bf{1-shot}               & \bf{5-shot}          \\ \hline
		%\bf{Pixel-nearest-neighbor} & 23.08\%                  & 24.70\%    \\
		\bf{Matching Network}                   & 45.00\% & 57.08\% \\
		\bf{Matching Network (FCE)}             & \bf{46.01}\%               & 57.38\%     \\
		%\bf{Prototypical}				& 0.00\%	&  0.00\% \\
		\bf{Meta-Learner LSTM}              &  38.37\%     & 53.65\% \\
		\bf{MAML}							&  31.52\%   & \bf{59.66}\% \\ \hline
		\bf{\shortname}                          &  39.10\%                    &   54.45\%             \\ \hline
	\end{tabular}
	\caption{Average test set classification accuracy on \emph{mini}DogsNet.}
  \label{dog-imagenet-results}
\end{table*}
%\end{document}

While the CUB results are promising, it is important to test the model's fine-grained abilities on a variety of data. For an alternate dataset we created a \emph{mini}DogsNet task, mirroring the \emph{mini}ImageNet dataset proposed in~\citet{matchingnets}. This dataset consists entirely of images from the ImageNet dog categories listed in that paper. We randomly selected 100 of those classes and used the same 64/16/20 random class split for training, validation, and testing\footnote{We note that this differs from the $L_{dogs}$ task described in that paper, which involved training on non-dog classes and testing on dog-specific classes.}.

%An alternative test of model generalization is the \emph{mini}DogsNet task first established in~\cite{matchingnets}. In this case, 118 classes are used for testing, but rather than being randomly sampled they include all classes of dogs. This is a more difficult task because it requires the models to generalize to an out-of-domain set of categories. As with \emph{mini}ImageNet, we use the same class splits as in~\citet{metalstm} but sampling our own images randomly when generating few-shot trials. We maintain the 84x84 image sizing and normalization procedure as in other experiments. We set the depth of blocks within the relational and comparison stages to 4 as in the \emph{mini}ImageNet experiments.
Results for the \emph{mini}DogsNet task are presented in Table~\ref{dog-imagenet-results}. The fine-grained task is more difficult than \emph{mini}ImageNet and performance is lower across the board. Whereas meta-learning approaches dominated the other baselines for broad classification, we find that this does not hold for \emph{mini}DogsNet. Matching networks with their full context embeddings is the highest performing baseline on the 1-shot experiments (46.01\%), while MAML still outperforms on the 5-shot (59.66\%).

Looking at the \shortname results, we again find that we are able to compete with these state-of-the-art approaches. In the 1-shot case we achieve 39.10\% on the 1-shot task and 54.45\% on the 5-shot. For 1-shot learning, this places us above either meta-learning baseline, but below matching networks.

%\subsection{Omniglot}
%\hilliard{TODO: Scott (time permitting?)}

\section{Conclusion}

We have introduced a novel \longname architecture for few-shot learning and evaluated its effectiveness across three image datasets. Our architecture deviates from previous approaches both in that it replaces a pre-defined distance metric with a learnable classifier and in that it explicitly conditions class representations to take into account the query image.
We achieve state-of-the-art performance for the Caltech-UCSD Birds dataset on both 1- and 5-shot experiments and show that the ability to condition is responsible for approximately a 5 percentage point boost in performance on that dataset.
The success of our approach on this fine-grained task also raises questions as to whether previous metric-based approaches might benefit from decisions being made by learned classifiers rather than pre-defined metrics.

%  perform only slightly worse than state-of-the-art baselines on both \emph{mini}ImageNet and \emph{mini}DogsNet datasets.
%A crucial component of our model is the ability to condition few-shot class embeddings by taking into account the specific query image which we demonstrate on the birds dataset by selectively removing this portion of the model. Our findings warrant further investigation into how information about classes is routed within few-shot architectures.

%These experiments demonstrate the effectiveness of negative examples in few-shot learning. Although we make use of the same number of images per experimental trial, we are still able to outperform traditional few-shot systems, including more sophisticated meta-learning models. Negative examples appear to be most useful in classification tasks involving fine-grained distinctions, supporting the notion that negative examples help our model to better identify the extent of class boundaries. Finally, although we have chosen negative examples randomly, future work might investigate how negative examples might be chosen intelligently to improve performance, either with a human-in-the-loop or through techniques such as active learning.

\section*{Acknowledgments}
This work was funded by the U.S. Government.

\bibliography{icml18}
\bibliographystyle{icml2018}

\end{document}